\documentclass[sigconf]{acmart}
\usepackage{natbib}
\AtBeginDocument{%
  \providecommand\BibTeX{{%
    \normalfont B\kern-0.5em{\scshape i\kern-0.25em b}\kern-0.8em\TeX}}}

\setcopyright{acmlicensed}
\copyrightyear{2018}
\acmYear{2018}
\acmDOI{XXXXXXX.XXXXXXX}

\acmConference[KDD'24]{Make sure to enter the correct
  conference title from your rights confirmation emai}{June 03--05,
  2018}{Woodstock, NY}
\acmISBN{978-1-4503-XXXX-X/18/06}

\begin{document}

\title{Towards a Systematic Approach to Design New Ensemble Learning Algorithms}


\author{João Mendes-Moreira}
\orcid{0000-0002-4802-7558}
\affiliation{%
  \institution{Faculdade de Engenharia Universidade do Porto}
  \city{Porto}
  \country{Portugal}}
\affiliation{%
  \institution{LIAAD - INESC TEC}
  \city{Porto}
  \country{Portugal}
}
\email{jmoreira@fe.up.pt}

\author{Tiago Mendes-Neves}
\orcid{0000-0002-4802-7558}
\affiliation{%
  \institution{Faculdade de Engenharia Universidade do Porto}
  \city{Porto}
  \country{Portugal}
}
\email{tiago.neves@up.pt}

\begin{abstract}
Ensemble learning has been a focal point of machine learning research due to its potential to improve predictive performance. This study revisits the foundational work on ensemble error decomposition, historically confined to bias-variance-covariance analysis for regression problems since the 1990s. Recent advancements introduced a "unified theory of diversity," which proposes an innovative bias-variance-diversity decomposition framework. Leveraging this contemporary understanding, our research systematically explores the application of this decomposition to guide the creation of new ensemble learning algorithms. Focusing on regression tasks, we employ neural networks as base learners to investigate the practical implications of this theoretical framework. This approach used 7 simple ensemble methods, we name them strategies, for neural networks that were used to generate 21 new ensemble algorithms. Among these, most of the methods aggregated with the snapshot strategy, one of the 7 strategies used, showcase superior predictive performance across diverse datasets w.r.t. the Friedman rank test with the Conover post-hoc test. Our systematic design approach contributes a suite of effective new algorithms and establishes a structured pathway for future ensemble learning algorithm development.

\end{abstract}

\ccsdesc[500]{Computing methodologies~Ensemble methods}



\keywords{Ensemble Learning, Error Decomposition, Neural Networks, Regression}

\begin{teaserfigure}
  \includegraphics[width=\textwidth]{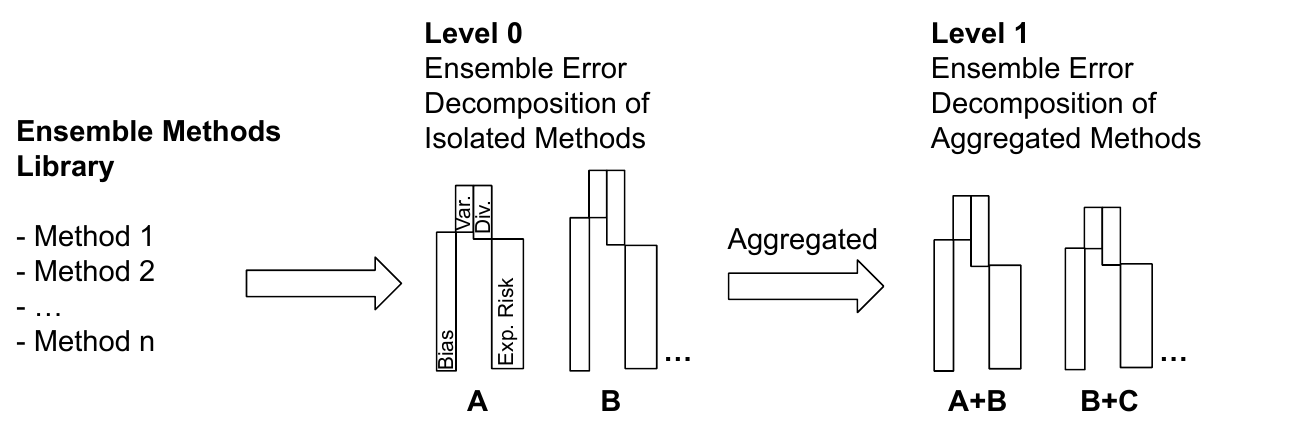}
  \caption{SA2DELA: Systematic Approach to Design Ensemble Learning Algorithms}
  \Description{SA2DELA: Systematic Approach to Design Ensemble Learning Algorithms.}
  \label{fig:teaser}
\end{teaserfigure}

\received{20 February 2007}
\received[revised]{12 March 2009}
\received[accepted]{5 June 2009}

\maketitle

\section{Introduction}
Ensemble Learning (EL) is a well-succeeded family of predictive algorithms that use several models instead of a single one to predict the target value of unlabeled instances. The most well-succeeded EL algorithms are homogeneous, i.e., they use the same method to generate the different models of the ensemble. EL methods can also be categorized as parallel or sequential. Parallel ensemble methods can train the models of the ensemble in parallel. When the models must be trained in sequence the ensemble method is named sequential. EL has two, optionally three, phases \cite{Mendes-Moreira2012}: generation, pruning (optional), and integration. In the generation phase, the set of models that will constitute the ensemble is generated. Optionally, in the pruning phase, a subset of them is selected. Then, the integration phase combines the predictions of the models from the ensemble to obtain the final ensemble prediction. In this paper, we only discuss homogeneous ensembles and we assume, without loss of generalization, two phases: generation and integration. 

Since the nineties, several studies have been proposed to decompose the ensemble error. Recently a new ensemble error decomposition was proposed \cite{Wood2023}. This new decomposition can be applied to both classification and regression error metrics, which happens for the first time. Previous decompositions could be applied only in the regression setting. 

In the last 30 years, many different homogeneous ensembles have been proposed. The use of the ensemble error decomposition is not, according to our knowledge, used to motivate the design of such algorithms despite in some of the cases there was a motivation to increase diversity or reduce bias, for instance. This study was motivated by 3 papers. One proposes the unified theory of diversity \cite{Wood2023}, the bias-variance-diversity decomposition of the ensemble error. The other two are Breiman's paper on random forests \cite{Breiman2001} and Webb's paper on multiboosting \cite{Webb2000} where the authors present well-succeeded algorithms by aggregating different strategies to generate the models of the ensemble. We name strategy to a portion of an algorithm that can be isolated and used to generate different models at each run. 

The use of the Ensemble error decomposition as a starting point to create new ensemble methods is proposed, consisting of aggregating in a unique algorithm two different strategies that can have complementary effects on the ensemble error. 

We use 7 different strategies to generate 21 new EL algorithms. The best overall algorithms are obtained by one of the strategies with some of the other strategies. All the experiments use the backpropagation algorithm in a multi-layer neural network to generate the ensemble models.

The main claimed scientific contributions are:
\begin{itemize}
    \item to use an ensemble error decomposition to combine the most complementary strategies in a single ensemble method;
    \item new algorithms with competitive results in the experiments using 21 new EL algorithms, 8 state-of-the-art ensemble methods, and a simple neural network, with one of them, dropout-snapshot, showing a better performance than its two constituent strategies.
\end{itemize}


Section \ref{SRW} reviews related work on ensemble error decomposition, ensemble methods for neural networks, and EL algorithms that result from aggregating different strategies to generate the base learners. Then, in Section \ref{SSA2DELA} the Systematic Approach to Design Ensemble Learning Algorithms (SA2DELA) is described. It is a 2-level approach. Sections \ref{SL0} and \ref{SL1} present the details of the 2 levels of the experiments with the SA2DELA. The statistical validation of the results is presented in Section \ref{SSV} and the conclusion and future works are discussed in Section \ref{SC}.

\section{Related Work}
\label{SRW}
\subsection{Ensemble error's decomposition}
There are three main formulas for ensemble generalization error decomposition in regression. First, Krogh \& Vedelsby \cite{Krogh1995} decomposes the error metric in bias and variance, the second by Ueda \& Nakano \cite{Ueda1996} decomposes it in bias, variance, and covariance, and the third, by Wood et al. \cite{Wood2023}, decomposes it in bias, variance, and diversity. This last approach can be used both for regression and classification problems. It is the one employed in this work despite we only study the regression setting.



However, there is one main limitation in all three error decomposition methods. They assume the use of Simple Averaging as the prediction integration strategy \cite{Krogh1995, Ueda1996, Wood2023}. This assumption expectedly skews the bias, variance, and diversity distribution when applying Weighted Averaging. Deriving this decomposition for Weighted Averaging would be more challenging or impossible due to the estimation of the ensemble error decomposition before the learning process, where, typically, the ensemble learns the weights.

\subsection{Neural network ensembles}
We survey ensemble methods that can also use neural network methods as base learners as well as ensemble methods that were specifically designed for neural networks. We start with the former.

Varying each estimator's training data provides a different framing of the problem. Despite Bootstrapping not being an ensemble-specific strategy, it may be utilized by repeatedly sampling with replacement dataset instances. Also, Pasting uses a smaller dataset, without demanding contiguous data samples, for each estimator \cite{Breiman1999}, Random Subspace employs a random input feature subspace policy on each estimator's dataset \cite{Ho1998}. This work uses decision trees. Later, a version for neural networks was presented under the name of Neural Random Subspace \cite{Cao2021}. Random Splits repeatedly sample from a dataset with a random data split in both train and test sets, with contiguous data samples, for each weak learner. K-fold Cross-Training splits the dataset into $k$ equally sized folds, feeds each estimator a different set $k-1$ folds, and may test it on the remaining holdout fold \cite{Krogh1995}. Note that $k$ is the number of estimators.

Bagging is an ensemble algorithm that uses bootstrap samples to fit independent parallel base learners \cite{Breiman1996}. There are multiple Bagging-based architectures but almost all are designed for decision trees as is the case of random forest \cite{Breiman2001}.

Boosting is, primarily, a bias reduction ensemble algorithm that builds upon prior chain models, fixes current prediction errors through attention refocusing (updates the dataset), and learns how to optimize each model's advantages. As a result, it turns weak learners into strong learners. There are multiple Boosting-based approaches such as AdaBoost, which at each iteration solves a "local" optimization problem and assigns weights to the data points and estimators based on their shown ensemble error contribution and performance respectively \cite{Freund1995}. Despite being developed for classification problems, Drucker \cite{Drucker1997} proposed a regression version named AdaBoost.R2. AdaBoost is expected to work well with unstable algorithms, such as decision trees or neural networks. Similar to Bagging, multiple boosting-based architectures were developed but, all of them, according to our knowledge, use decision trees as base learners.

Stacked generalization, also named stacking, is an ensemble technique with two levels: in the first level, a set of models is generated using the same or different base learners, and in the second level, another learner is used to learn the weights for the weighted average \cite{Wolpert1992}. Any learner can be used in both levels of stacking including neural networks.

Now, ensemble approaches specifically designed for using neural networks as base learners are presented. Snapshot generates an ensemble with estimators that visited multiple local minima but not necessarily from contiguous training epochs \cite{Huang2017}. Stochastic Gradient Descent with Warm Restarts (SGDR) promotes even greater Snapshot diversity. This approach aggressively cycles the learning rate, thus avoiding individual estimators getting stuck in the same local minima \cite{Huang2017}. Another alternative that stems from Snapshot is Polyak Averaging, which averages into a single Network multiple sets of noisy weights from contiguous training epochs close to the end of a training run \cite{Polyak1992}. 

Negative Correlation Learning, motivated by the works of Naonori Ueda and Ryohei Nakano \cite{Ueda1996}, is a strategy to promote mutual model diversity and lower the correlation between base learners' predictions. When generating a model for the ensemble, an added penalty term to the Neural Network's objective function promotes a negative correlation between the new model and the previously generated models \cite{Rosen1996, Liu1999}.

Despite Dropout not being an ensemble-specific strategy, it can be used as such. It promotes diversity during the learning process of each Neural Network in the ensemble by randomly dropping a given percentage of nodes \cite{Srivastava2014}.

There are various options for varying the base models' prediction integration strategy. Simple Averaging combines independent base learner's predictions with equally distributed weights \cite{Clemen1989}. Simple Averaging of models extracted from contiguous training epochs close to the end of a training run is defined as Horizontal Averaging \cite{Xie2013}. Still, it might be helpful to consider each respective base learner's demonstrated accuracy in determining the final result, thus using a Weighted Average.

\subsection{Combining strategies to generate ensembles}

Each EL algorithm has its strategy/strategies to generate the models that will constitute the ensemble. The existing approaches that combine strategies to obtain ensembles with better predictive performance use decision trees as base learners. We describe them because their principles also motivate the design of new neural network ensembles by combining strategies. 

An example of a strategy used for unstable algorithms (e.g.: decision trees, artificial neural networks) is the use of bootstrapping to select the training sets used to train each of the models as it is done in bagging \cite{Breiman1996} and random forest \cite{Breiman2001}. In the last case, random forests, an additional strategy is used: each split, in the decision tree training, is chosen from a randomly selected subset of features \cite{Ho1998}. The number of features selected is a hyperparameter of random forests. It seems that Leo Breiman was trying to increase the diversity of the trees.

Some other proposed approaches exist to combine strategies for generating ensembles, each offering different breakthroughs from those presented in this work. MultiBoosting combines AdaBoost with wagging, a variant of Bagging using C4.5 as the base learners achieving better results and execution time than the constituent algorithms \cite{Webb2000}. Multistrategy Ensemble Learning investigates the hypothesis that accuracy improvement is due to base learners' increased diversity. So three new multistrategy Ensemble Learning techniques were developed with results showing they are, on average, more accurate than their base strategies \cite{Webb2004}. Another work named Random Patches merges Random Subspace with Pasting \cite{Louppe2012}.

\section{SA2DELA: A systematic approach to design ensemble learning algorithms}
\label{SSA2DELA}

SA2DELA assumes that a set of ensemble strategies was previously chosen. It is advisable to consider simple ensemble strategies. For example, we use in the experiments the negative correlation learning as proposed by Rosen \cite{Rosen1996} instead of subsequent methods also based on negative correlation learning as is the case of \cite{Liu1999}. The reason is that simpler approaches are easier to aggregate with other simpler approaches. This is also the reason why we do not use the AdaBoost algorithm \cite{Freund1995} because it is a method that is difficult to aggregate with other ensemble methods. Moreover, the chosen strategies have meaningful differences between them. Strategies with small variations to other ones already chosen were avoided.

The 2-levels of SA2DELA are: 
\begin{enumerate}
\item {level-0}: it decomposes the ensemble error in bias, variance, and diversity \cite{Wood2023} for each of the ensemble strategies under study.
\item{level-1}: it combines pairs of strategies to create a multitude of new ensemble algorithms. The choice of the pairs to combine is an option of the researcher but the information on the ensemble error decomposition can give insights on the pairs to combine.
\end{enumerate}

\section{level-0 experiments}
\label{SL0}

This work only uses the neural network method as the base learner and assumes the regression setting. 

\subsection{Experimental setup}
\subsubsection{Data Acquisition and Preprocessing}
The experimental setup involved acquiring and preprocessing datasets from the OpenML suite, specifically the benchmark suite OpenML-CTR23 \cite{Fischer2023}. This benchmark suite comprises diverse tasks, facilitating a comprehensive evaluation across different problem domains. The data was fetched using the OpenML Python API. The preprocessing steps were the following:

\begin{itemize}
\item Categorical Feature Encoding: Categorical features were identified and encoded using a Leave-One-Out Encoder. This encoder was chosen for its effectiveness in handling categorical variables \cite{Seca2021}.
\item Data Cleaning and Normalization: The datasets were cleaned by dropping features with missing values. Following this, feature normalization was conducted by subtracting the mean and dividing by the standard deviation of each feature, ensuring a common scale across all features.
\item Handling Zero Variance Columns: Columns with zero variance were removed, as they do not contribute to model learning.
\end{itemize}

This preprocessing pipeline was applied uniformly across all datasets in the suite, ensuring a consistent data format and structure for subsequent modeling and analysis. 

\subsubsection{Hyperparameter Optimization}
To reduce the impact of hyperparameters on the performance of the ensembles, we searched for four prototypes of neural network architectures that could solve the suite of problems. We selected four prototypes since it allowed enough diversity for each dataset to have a good-performing base learner while keeping the search for the hyperparameters for the base learners at a low cost.

The Optuna framework \cite{Akiba2019} for hyperparameter optimization was employed, focusing on a range of architectures with varying hidden layer sizes, activation functions, learning rates, and fixed epoch counts.  Each dataset was divided using a 3-fold cross-validation scheme with shuffling enabled, and the optimization was performed intra-fold. The random state was set to 42, ensuring the folds are equal in the optimization and testing processes, avoiding mixing training and testing data from one process to another.

Each model configuration varied in:
\begin{itemize}
\item  Number of units in two hidden layers, selected from \\$\{2^4, 2^5, 2^6, 2^7, 2^8\}$. 
\item Activation function, chosen from \{'relu’, ‘sigmoid’, ‘tanh’\}. 
\item Learning rate, within the range $[1e^{-3}, 1e^{-1}]$. 
\item Fixed epoch count of 10, previously determined empirically.
\end{itemize}

Each model was trained using a Mean Squared Error (MSE) loss function. A cosine annealing learning rate strategy was applied, updating the learning rate at each epoch based on the model's initial learning rate and epoch count. Validation loss was calculated post-training to assess model performance.

Optuna's hyperparameter optimization was executed over 100 trials. Each trial involved training and evaluating all model configurations across all folds of each dataset. The model with the lowest validation loss was selected for each fold. The optimization goal was to minimize the global loss, defined as the sum of the minimum validation losses across all folds and datasets. Table \ref{table:NN_arq} presents the four architectures selected.

\begin{table}
\captionsetup{justification=centering}
\caption[The four architectures selected]{The four architectures selected where SHL stands for Size Hidden Layer, AF for Activation Function, LR for Learning Rate, and Sb for Selected by}
\label{table:NN_arq}
\centering
\vspace{3mm}
\begin{tabular}{cccccc}
ID & SHL 1 & SHL 2 & AF & LR & Sb \\ \hline
M0 & 256 & 16 & ReLU & 0.02024 & 39 \\
M1 & 16 & 32 & tanh & 0.06929 & 9 \\
M2 & 128 & 64 & ReLU & 0.03037 & 50 \\
M3 & 32 & 32 & tanh & 0.09489 & 7 \\ \hline 
\end{tabular}
\vspace{-4mm}
\end{table}

\subsubsection{Ensemble Testing}
The ensembles were constructed using a custom class to manage the base learners and handle data sampling. Various ensemble methods were tested, including:
\begin{itemize}
\item Single Model: A single estimator.
\item Simple Average: An ensemble averaging the outputs of multiple estimators.
\item Bagging: Utilizing bootstrapping to create diverse training sets, one per estimator.
\item Random Subspaces: Each estimator is trained on a random subset of features. We set the parameter max\_features=0.7.
\item Pasting: Similar to bagging but without replacement in sampling. We set the parameter max\_samples=0.7.
\item Dropout: Incorporating dropout rates in the base learners. We set the parameter dropout\_rate=0.2 for each hidden layer.
\item Snapshot: Capturing snapshots of base learners at different epochs. We use the Cyclic Cosine Annealing Learning Rate, which returns the learning rate to the initial value every 10 epochs. Furthermore, instead of training different base learners, we create the base learners by making a copy of the model every 10 epochs, before the learning rate resets.
\item Negative Correlation Learning (NCL): Employing a custom loss function to encourage diversity among the learners. We set the parameter lambda=0.1.
\item Stacking: Using predictions of base learners as input to a second-level learner. The second-level learner is another Neural Network with a single hidden layer of size equal to the number of base learners. The activation function is ReLU, and the learning rate is 0.02.
\end{itemize}

Each ensemble was trained multiple times (10 iterations) on datasets partitioned into three folds (3-fold cross-validation), the same as the optimization process. The base learner architecture is randomly selected from the two folds used in training. To evaluate the bias-variance-diversity decomposition, we use the Decompose library \cite{Wood2023}.

The experimental environment ran on Pytorch. The code is available at \url{https://github.com/nvsclub/EnsemblingNeuralNetworks}.

\subsection{Strategies to generate the ensemble models}
As previously said, ensemble methods that use simple strategies to generate the models are used. The seven neural network ensemble methods used in level-0 are described in Table \ref{table:generation_integration_strategies}.

\begin{table}
\captionsetup{justification=centering}
\caption[Overview of Generation Mode/Integration Methods]{\centering Overview of Generation Mode (GEN), Integration Method (INT), and main reference (REF) for the main neural network ensemble methods for regression. Par stands for Parallel, Str for Stream (means that we generate the base models from the same original base learner), Seq for Sequential, SA for Simple Average, and WA for Weighted Average. The Random Subspace is our version of the \cite{Ho1998} work where each neural network model is trained using a random subset of the predictive features.}
\label{table:generation_integration_strategies}
\centering
\vspace{3mm}
\begin{tabular}{cccc}
\multicolumn{1}{c}{Ensemble Algorithm} & GEN & INT & REF \\ \hline
Random Subspace & Par & SA & \cite{Ho1998} \\ 
Pasting & Par & SA & \cite{Breiman1999} \\ 
Snapshot & Str & SA & \cite{Huang2017}\\ 
Negative Correlation Learning & Seq & SA & \cite{Rosen1996}\\
Dropout & Par & SA & \cite{Srivastava2014}\\ 
Bagging & Par & SA & \cite{Breiman1996} \\ 
Stacking & Par & WA & \cite{Wolpert1992} \\ \hline
\end{tabular}
\vspace{-4mm}
\end{table}

\subsection{Results and discussion}

\begin{figure}
\includegraphics[width=\columnwidth]{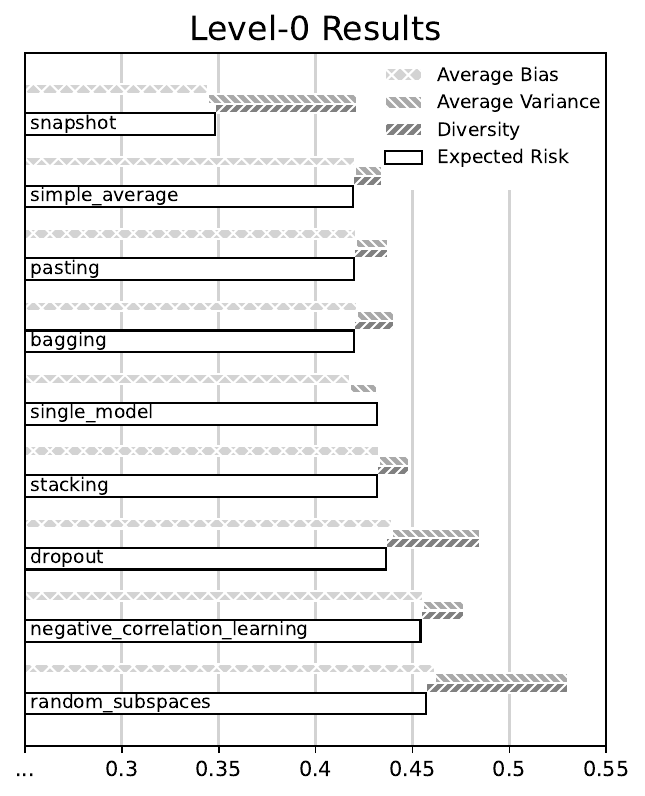}
\caption{Level-0 results. Neural Network ensemble results. Besides the 7 algorithms described, an ensemble method using the simple average as the integration method and a single neural network were also used as baselines.} 
\label{fig:level_0_results}
\end{figure}

The results of the level-0 experiments are shown in Figure \ref{fig:level_0_results}. 
 
The results are presented structured according to the bias-variance-diversity decomposition, i.e., the average bias and the average variance sum up while the diversity subtracts resulting in the expected risk. The methods are ranked by the increasing order of the expected risk.
Four groups can be identified, the first composed by the snapshot method, the second constituted by the simple average, pasting, and bagging, the third group with the single model, stacking, and dropout, and finally, the fourth group with the negative correlation learning and the random subspaces, It is also noticeable that the largest positive difference between diversity and the average variance is obtained by the random subspaces and dropout. The snapshot presents a large difference from all others both in terms of the expected risk and the average bias.

\section{Level-1 Experiments}
\label{SL1}

Level-1 can use the information obtained in level-0 to choose the most promising aggregation pairs. From level-0, the snapshot due to its low bias, and dropout and random subspace due to their larger positive difference between diversity and average variance, seems to be the most promising. However, in this study, we opt to do all possible combinations of pairs resulting in $C_2^7 = 21$ pairs allowing us to better understand this approach. All these 21 algorithms are new despite one of them, the aggregation between bagging and pasting can be seen as a small variation of random patches \cite{Louppe2012}. 

\subsection{Experimental Setup}
The experimental setup used in level-1 experiments is equal to the experimental setup used in level-0 experiments. 

\subsection{Results and discussion}

\begin{figure}
\includegraphics[width=\columnwidth]{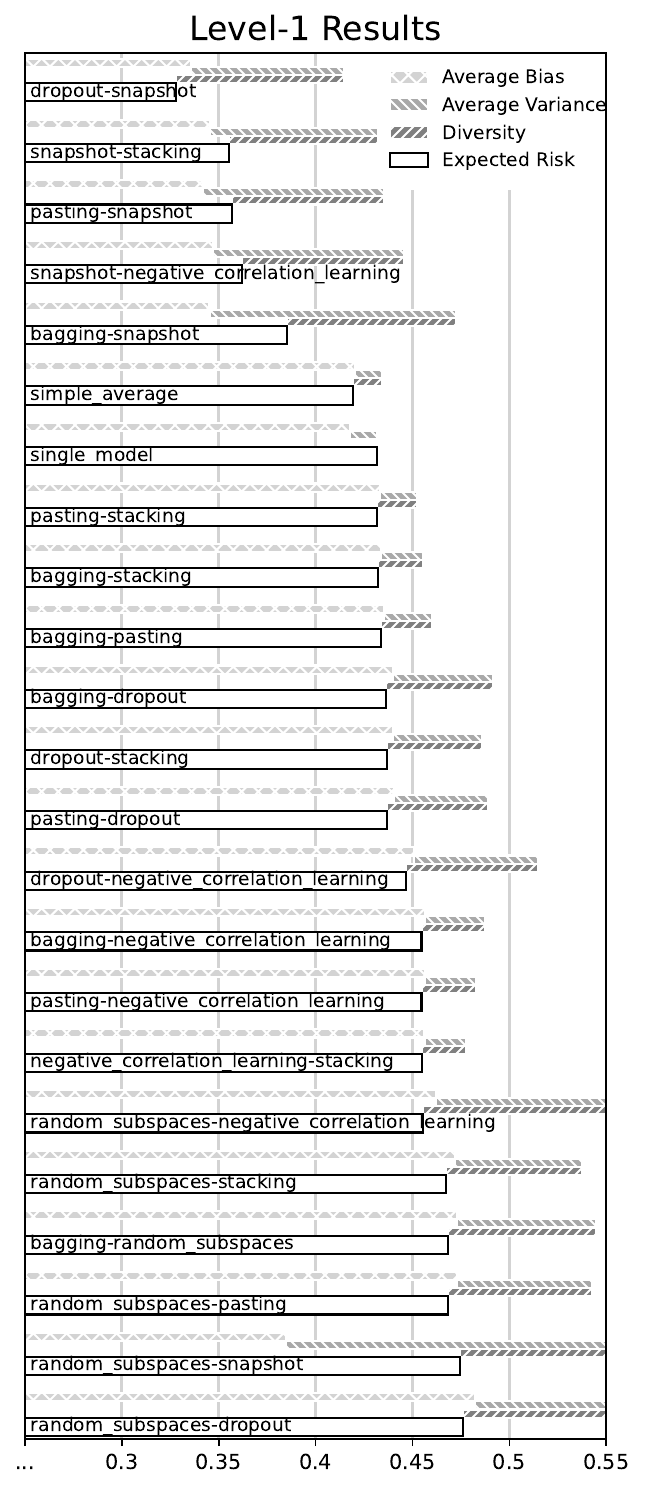}
\caption{Level-1 results showing the 21 new ensemble methods plus the simple average ensemble and the single neural network model.} 
\label{fig:level_1_results}
\end{figure}

The results of the level-1 experiments are shown in Figure \ref{fig:level_1_results}. 

The results show that the snapshot method aggregated with any other method except the random subspace gets better results than any of these other methods alone. However, the unique aggregated method that beats the snapshot is the aggregation of the snapshot with dropout. All aggregations of random subspace are the worst performance aggregations. The second worst is the aggregations of the negative correlation learning except the one with snapshot. The aggregations between the remaining methods, bagging, pasting, stacking,  and dropout, get results between these extremes. Moreover, it is interesting to observe that dropout and random subspace were the ones with a larger positive difference between the diversity and the average variance in level-0. Despite that, the aggregation of snapshot with dropout is the best overall aggregation while the aggregation between snapshot and random subspace is the second worst overall aggregation in level-1. The bad result of the aggregation between snapshot and random subspace is mainly due to the largest negative difference between the diversity and the average variance. These results show that despite the insights that can be obtained by analysing the level-0 results, the results of the aggregations do not depend solely on the level-0 results of the aggregation constituents. It is also interesting to observe that the snapshot aggregated with random subspace gets a larger average bias than the snapshot alone, and aggregated with dropout gets a lower average bias than the snapshot alone, and that aggregated with any other strategy except dropout gets a higher negative difference between the diversity and the variance than the snapshot alone.

\subsection{Sensitivity to the ensemble size}

\begin{figure}
\includegraphics[width=\columnwidth]{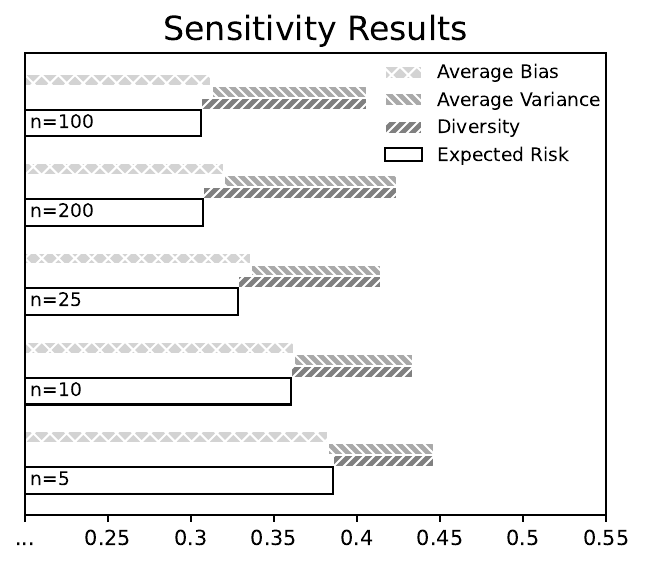}
\caption{Ensemble size sensitivity for the dropout-snapshot aggregated method ordered by the increasing order of the expected risk.} 
\label{fig:ensembleSize}
\end{figure}

All the experiments shown in the previous sections use ensembles with 25 models. Naturally, there is no guarantee that this is the best size for the ensembles. In Figure \ref{fig:ensembleSize} the results obtained with 5, 10, 25, 100, and 200 models using the dropout-snapshot aggregated method are shown. 

These results follow \cite{Wood2023} where the authors show that by increasing the number of models in the ensemble, both the average bias and the average variance kept constant while the diversity increased resulting in a decrease in the expected error. It can be seen that using around 100 models allows for reducing the expected error.

\section{Statistical validation}
\label{SSV}

The results of both level-0 and level-1 were statistically validated using the Friedman rank test \cite{Demsar2006} to evaluate the existence of differences between the results of the different methods, and the Conover post-hoc test \cite{Conover1999} to evaluate which pairs of methods are statistically different. All tests used a significance level of 5\%.

For level-0, the p-value of the Friedman rank test was $7.28e^{-18}$. For level-1, this value was $1.04e^{-51}$.

\begin{figure*}
\includegraphics[width=2\columnwidth]{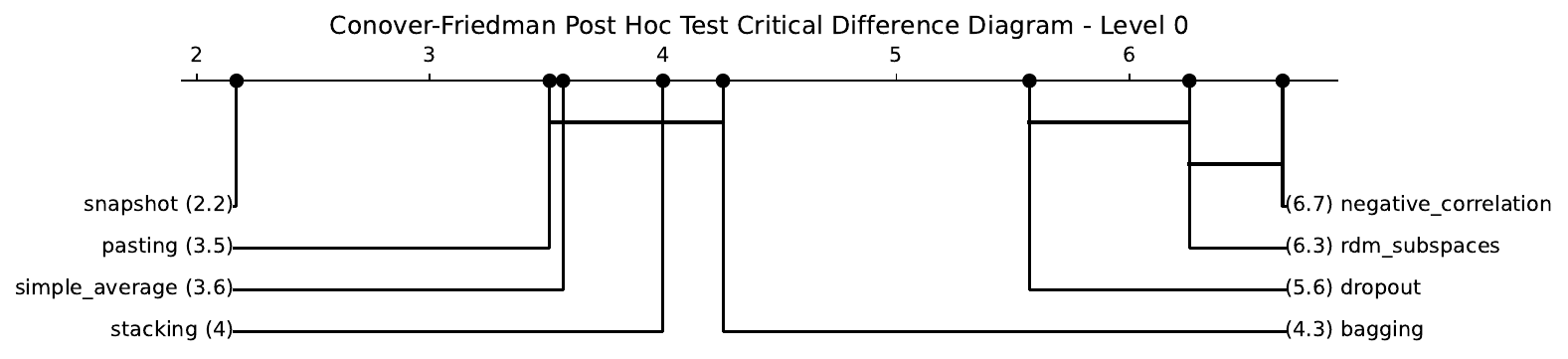}
\caption{The Friedman-Conover post-hoc test for level-0 experiments.} 
\label{fig:conover0}
\end{figure*}

\begin{figure*}
\includegraphics[width=2\columnwidth]{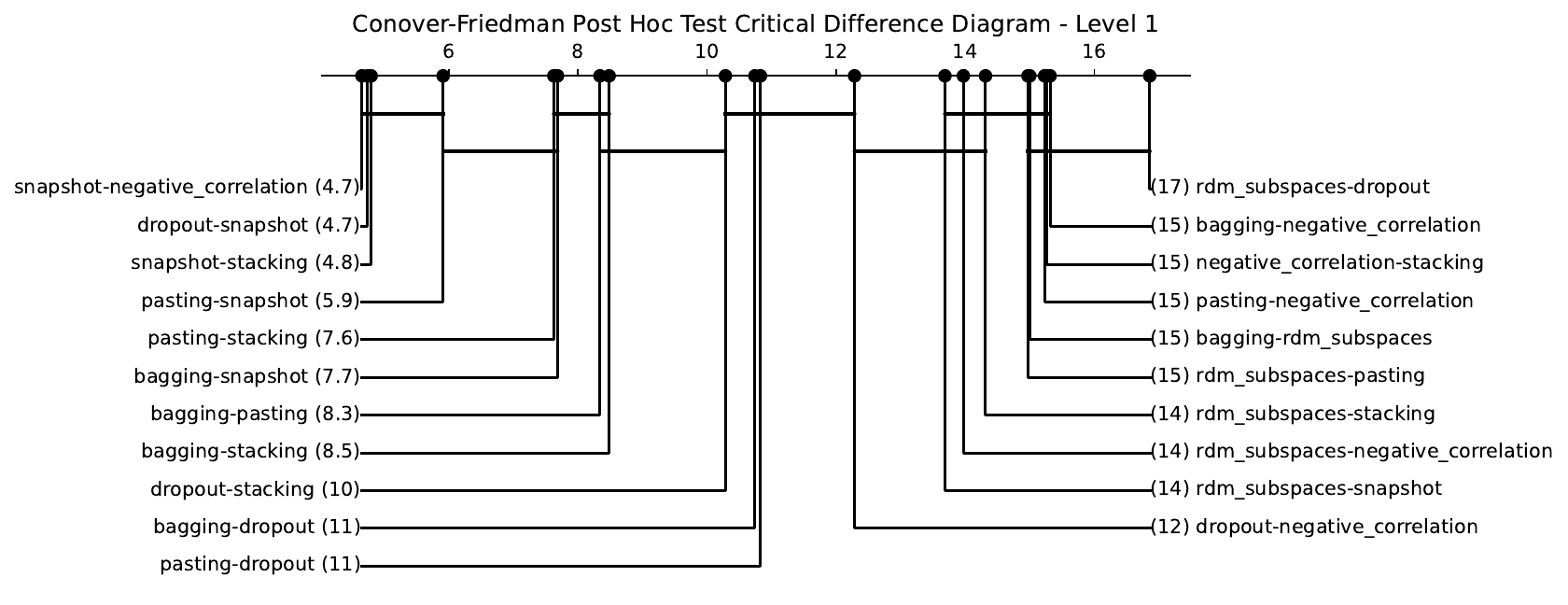}
\caption{The Friedman-Conover post-hoc test for level-1 experiments.} 
\label{fig:conover1}
\end{figure*}

The Conover-Friedman post-hoc distances for level-0 and level-1 are shown respectively in figures \ref{fig:conover0} and \ref{fig:conover1}.

The results seem to be different from the results presented in Figures \ref{fig:level_0_results} and \ref{fig:level_1_results}. The results from Figures \ref{fig:level_0_results} and \ref{fig:level_1_results} were obtained by averaging the standardized root mean squared error values while the statistical validation used the average of the ranks obtained by each method in the 35 datasets. The observed results differences are due to the differences of these two metrics.

\section{Conclusions}
\label{SC}

The ensemble decomposition can help design better ensemble algorithms. The SA2DELA is a fully empirical 'process' inspired by the random forest method and ensemble error decomposition methods aiming to systematize the design of new ensemble algorithms in an informed way. A better understanding of the weaknesses and strengths of the existing ensemble methods can be obtained through the ensemble error decomposition. Despite it is not possible to fully estimate the result of aggregating two ensemble strategies, this framework gives insights into the behaviour of the different ensemble strategies. This study focused on neural network ensembles for regression. New promising neural network ensemble algorithms, namely the ones that aggregates snapshot with the negative correlation learning, dropout or stacking are the most promising ones.

The same approach can be applied to classification or using another base learner, such as decision trees.

\begin{acks}
This work is financed by National Funds through the Portuguese funding agency, FCT - Fundação para a Ciência e a Tecnologia, within project UIDB/50014/2020.
\end{acks}

\bibliography{mybibliography}

\end{document}